\documentclass{article}
\usepackage{spconf,amsmath,graphicx,cite}
\usepackage{siunitx,wrapfig}
\usepackage{todonotes}
\usepackage{booktabs}
\usepackage{multirow}
\usepackage{multicol}
\usepackage{amssymb}
\usepackage{cite}
\usepackage{array}
\usepackage{fancyhdr}
\begin{document}
% \title{The Cost of Speaking: Emission-Aware Reward Design for Proactive Video Assistants}
\title{\textbf{SHUSH}: Emission-Aware Reward Design for Proactive Video Assistants\vspace{-3mm}}

\name{\begin{tabular}{c}Shawn Liang$^1$ \ \ Soamar Homsi$^2$ \ \ Gourav Datta$^1$\end{tabular}\vspace{-3mm}}
\address{
\shortstack{
$^{1}$Case Western Reserve University, Cleveland, OH, USA\\
$^{2}$Air Force Research Laboratory, Rome, NY, USA
}
}
%\address{$^1$Case Western Reserve University, Cleveland, OH, USA \ $^2$Air Force Research Laboratory, Rome, NY, USA}
%
% For example:
% ------------
%\address{School\\
%	Department\\
%	Address}
%
% Two addresses (uncomment and modify for two-address case).
% ----------------------------------------------------------
%\twoauthors
%  {A. Author-one, B. Author-two\sthanks{Thanks to XYZ agency for funding.}}
%	{School A-B\\
%	Department A-B\\
%	Address A-B}
%  {C. Author-three, D. Author-four\sthanks{The fourth author performed the work
%	while at ...}}
%	{School C-D\\
%	Department C-D\\
%	Address C-D}
%

%\ninept
%
\maketitle
%
%\vspace{-3mm}
\begin{abstract}
% Proactive video assistants must decide not only what to say, but also when to speak. Existing reinforcement-learning objectives can reward timely, correct responses without adequately accounting for response volume, thereby encouraging unnecessary or repeated emissions. We identify this failure mode in MMDuet2, whose proactive area-under-the-curve (PAUC) reward promotes timely responses but does not explicitly price the number of emissions. We show analytically that this can make additional responses weakly penalized or effectively free, systematically encouraging over-emission. We then propose an emission-aware F1 reward for interval-valued annotations, using exclusive span--emission matching to price false emissions and missed spans. Under the same training recipe, our policy reduces replies on the visual subset of OmniPro from 6,961 to 1,216 (82.5\%), increases precision from 0.261 to 0.448, and maintains comparable F1 (0.366 vs.\ 0.375) score. ProactiveVideoQA results further show that PAUC can favor high-emission policies because it is insensitive to response count. Our results establish emission pricing as a key principle for proactive video assistants.

Proactive video assistants must decide not only what to say, but also when to speak. Existing reinforcement-learning objectives can reward timely, correct responses without accounting for response volume, thereby encouraging unnecessary or repeated emissions. We identify this failure mode in MMDuet2, whose proactive area-under-the-curve (PAUC) reward promotes timely responses but does not price the number of emissions. We show analytically that this makes additional responses weakly penalized or effectively free, systematically encouraging over-emission. We then propose an emission-aware F1 reward for interval-valued annotations, using exclusive span--emission matching to price false emissions and missed spans. Under MMDuet2's training recipe, our policy on the visual subset of OmniPro replies 2.5$\times$ less often than MMDuet2 while raising timing F1 from 0.259 to 0.396 and content-aware F1 from 0.103 to 0.154; a more conservative variant cuts replies by 82.5\% and raises precision from 0.161 to 0.418. On ProactiveVideoQA, the same reward adds replies where the baseline is nearly silent, and PAUC ranks a heavily over-emitting policy second despite near-last F1, because it is insensitive to response count. Our results indicate that proactive assistants should be trained and evaluated with objectives that price emissions.\vspace{-2mm}
\end{abstract}

%
%\begin{keywords}
%SNN, MAC, AC, accelerators, in-sensor
%\end{keywords}
%
\vspace{-4mm}
\section{Introduction}
\vspace{-2mm}
\label{sec:intro}
%Video large language models (Video-LLMs) are increasingly moving from offline video understanding toward continuous streaming interaction, where frames arrive sequentially and the model must decide autonomously when to respond~\cite{videollm-online,wang2025mmduet,wang2026mmduet2}. In such proactive systems, response timing is part of the prediction problem: speaking too late reduces utility, while speaking unnecessarily produces repetitive or unsupported responses. We refer to each model-generated reply at a decision turn as an \emph{emission}.

Video large language models (Video-LLMs) are increasingly moving from offline video understanding toward continuous streaming interaction, where frames arrive sequentially and the model must decide autonomously when to respond~\cite{videollm-online,wang2025mmduet,wang2026mmduet2}. In such proactive systems, response timing is part of the prediction problem: speaking too late reduces utility, while speaking unnecessarily produces repetitive or unsupported responses. Because the assistant initiates these interactions without being prompted, each unnecessary reply consumes user attention, can distract from genuinely relevant events, and adds avoidable inference and communication overhead; repeated low-value replies can also erode user trust. We refer to each model-generated reply at a decision turn as an \emph{emission}.

Recent systems have therefore begun learning when to speak rather than relying only on fixed inference thresholds. MMDuet2~\cite{wang2026mmduet2}, for example, formulates response triggering as a per-turn textual decision and optimizes the policy using reinforcement learning. Despite improved timing performance, recent evaluation shows that proactive Video-LLMs can still achieve high temporal scores by responding frequently, increasing recall at the expense of precision~\cite{li2026streampro}. This raises a fundamental question: \emph{does the learned policy over-emit because optimization fails, or because the reward itself makes frequent emission advantageous?}

\begin{figure}
    \centering
    \includegraphics[width=0.5\textwidth]{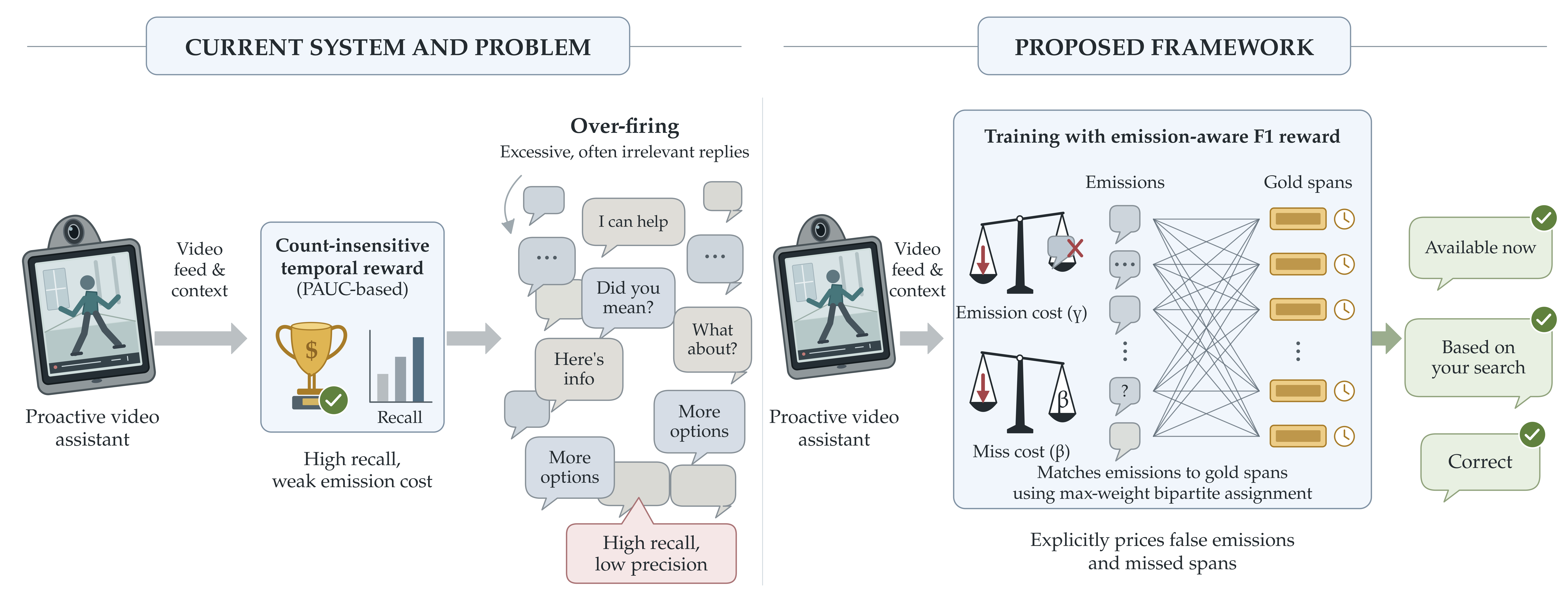}
    \vspace{-8mm}
    \caption{\footnotesize Overview of the proposed emission-aware reward design. Count-insensitive temporal rewards can favor repeated or unnecessary responses, whereas our interval-aware F1 formulation prices false emissions and missed spans through exclusive emission--span matching.}
    \label{fig:overview}
    \vspace{-6mm}
\end{figure}

We show that the latter can occur. In MMDuet2, the dominant Proactive Area Under the Curve (PAUC)-based reward integrates response quality over an annotated temporal span, rewarding early improvements but not explicitly accounting for the total number of emissions. An additional reply can therefore increase reward when it improves the running score, while costing nothing when it does not, and the auxiliary penalties for out-of-span and repetitive replies weaken as the number of decision turns grows. Section~\ref{sec:analysis} characterizes these incentives and shows that over-emission becomes increasingly favorable when relevant events are sparse relative to decision opportunities.

To address this mismatch, we propose an \emph{emission-aware F1 reward} for interval-valued annotations that explicitly couples reward to the number of responses produced by the model. Our formulation uses one-to-one maximum-weight matching between emissions and gold spans, allowing each span to credit at most one reply while explicitly penalizing unmatched emissions and missed spans. As a result, an additional response must provide sufficient value to justify its cost rather than serving as a nearly free additional attempt. Our contributions are threefold: (1) an analytical diagnosis of how PAUC-based reward design can systematically incentivize over-emission; (2) an interval-aware, emission-priced F1 reward based on exclusive span--emission matching; and (3) an empirical evaluation showing an 82.5\% reduction in emissions on OmniPro while raising precision from 0.161 to 0.418 and F1 from 0.259 to 0.346. Figure~\ref{fig:overview} contrasts the two reward structures.

\vspace{-3mm}
\section{Related Work}
\vspace{-1mm}
\label{sec:related}

\noindent\textbf{Proactive streaming Video-LLMs.}
Streaming Video-LLMs must determine both response content and response timing~\cite{videollm-online,yang2026towardonline}. Existing triggering mechanisms include token-driven decisions~\cite{videollm-online,yu2025eyeswideopen,chen2025livecc}, dedicated classification heads~\cite{wang2025mmduet,qian2025dispider,wang2025streambridge}, and heuristic visual or uncertainty criteria~\cite{yang2025livestar,yao2025timechatonline}. Recent methods increasingly learn this decision directly through reinforcement learning. MMDuet2~\cite{wang2026mmduet2}, for example, formulates triggering as a per-turn textual decision optimized via RL, eliminating the need for fixed inference thresholds. We focus on the reward design for such learned response policies.

\noindent\textbf{Timing metrics and rewards.}
StreamingBench~\cite{lin2024streamingbench} and OVO-Bench~\cite{niu2025ovobench} evaluate temporal correctness but score at most one response per query. ProactiveVideoQA~\cite{wang2025proactivevideoqa} introduces PAUC, which integrates response quality over an annotated temporal span but does not explicitly account for the number of emitted responses. More recent benchmarks address over-triggering directly: OmniPro~\cite{omnipro2026} penalizes unnecessary responses, while StreamPro-Bench~\cite{li2026streampro} evaluates response trajectories using precision-recall F1. Correspondingly, MMDuet2~\cite{wang2026mmduet2} optimizes a PAUC-derived reward that favors early, correct responses while adding penalties for repetition, out-of-span emissions, and verbose prefixes, whereas StreamPro~\cite{li2026streampro} introduces a count-sensitive turn-level F1 term for timestamp-valued targets. Both score response content with an LLM judge~\cite{li2024llmasjudges}. In contrast, we analyze why count-insensitive temporal objectives favor over-emission and propose an emission-aware formulation for interval-valued annotations.

\vspace{-3mm}
\section{When Temporal Rewards Encourage Over-Emission}
\vspace{-1mm}
\label{sec:analysis}

We use MMDuet2~\cite{wang2026mmduet2} as a representative case study
of RL-based proactive response timing. On the visual subset of OmniPro,
it produces 6,961 replies for 1,723 annotated triggers, motivating closer
inspection of the underlying objective. These excess emissions occur both
outside annotated events and as repeated responses within already-covered
spans. Rather than viewing this as a model-specific optimization failure,
we identify two broader structural incentives in count-insensitive temporal
objectives: additional predictions can carry little or no marginal cost,
and normalized penalties can weaken as the number of decision opportunities
increases.

\vspace{-3mm}
\subsection{Marginal incentives in a PAUC-based objective}
\vspace{-1mm}
\label{subsec:arch}

Let a video rollout consist of $K$ total decision turns, with $M$ turns
falling outside any annotated ground-truth span. The annotation provides
$N$ gold spans bounded by $[t^{start},t^{end}]$. Within a given span,
the model emits $P$ replies at timestamps $\tau_p$, each assigned a
content score $s_p\in[0,S]$ (with $S=4$) by an LLM judge. The published reward is
\begin{equation}
r = 3r_\mathrm{PAUC}
  + 2r_\mathrm{rep}
  + 0.5r_{\mathrm{in\_span}}
  + 2r_\mathrm{pfx}.
\label{eq:reward}
\end{equation}
\noindent The dominant term, $r_\mathrm{PAUC}$, is derived from the PAUC metric of ProactiveVideoQA~\cite{wang2025proactivevideoqa}.
It represents the response score as a step function over time, initialized
at a floor of $s_0=0.5$ at $t^{start}$, and measures its area as a
fraction of the maximum possible area:
\begin{equation}
\begin{split}
\mathrm{PAUC}
&= \frac{1}{(t^{end}{-}t^{start}){\times}S}
\Biggl[
(\tau_1{-}t^{start}){\times}0.5 \\
&\quad +
\sum_{p=1}^{P-1}(\tau_{p+1}{-}\tau_p){\times}s_p
+ (t^{end}{-}\tau_P){\times}s_P
\Biggr].
\end{split}
\label{eq:pauc}
\end{equation}

The reward averages PAUC over the $N$ spans. Following the official implementation, timestamps are evaluated at $\tilde{\tau}_p=t^{start}+(1-\phi)(\tau_p-t^{start})$ with $\phi=0.5$. The remaining terms penalize semantic repetition, out-of-span emissions, and repeated prefixes; importantly, these penalties are normalized respectively by emission count, the number of out-of-span turns $M$, and the total number of turns $K$.

Let $\xi\in[0,1]$ denote an emission's fractional offset within a span.
Adding a final reply with score $s$ when the running score is
$s_\mathrm{prev}$ yields
\begin{equation}
\Delta\mathrm{PAUC}
=
\bigl(1-(1-\phi)\xi\bigr)
\bigl(s-s_\mathrm{prev}\bigr)/S.
\label{eq:marginal}
\end{equation}

The marginal gain depends on $s-s_\mathrm{prev}$, rewarding only improvements over the current response. However, the emission count $P$ does not appear in the PAUC normalization, so additional attempts can increase reward without incurring cost when they fail to improve the running score.

\begin{figure}
    \centering
    \includegraphics[width=0.5\textwidth]{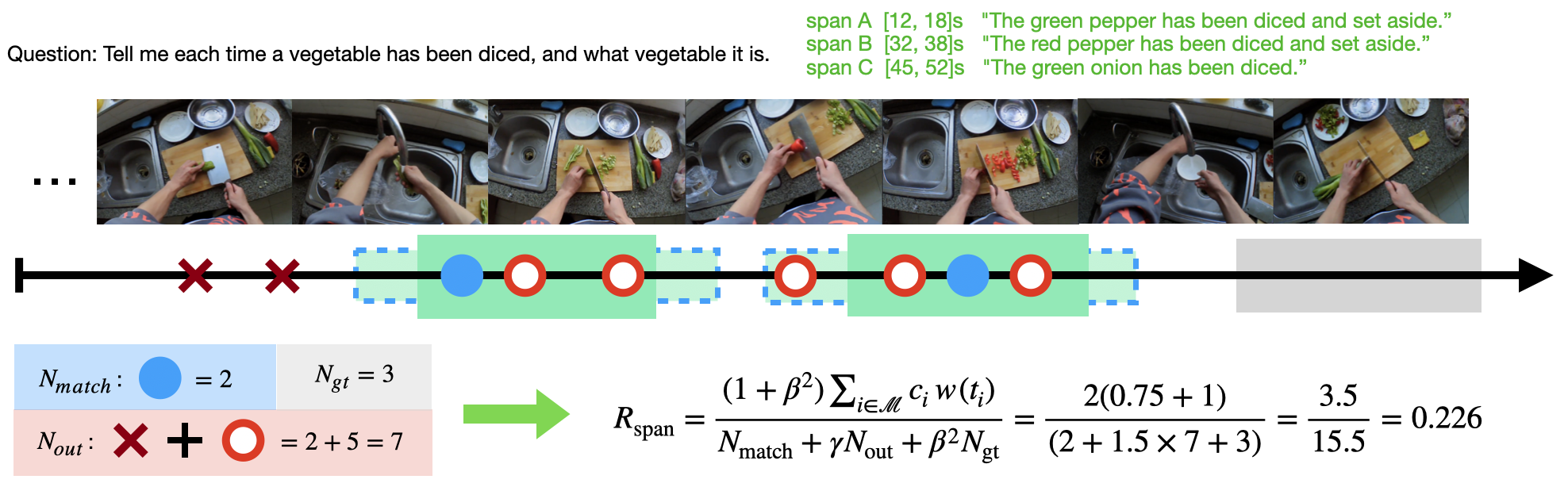}
    \vspace{-7mm}
    \caption{\footnotesize The emission-aware F1 reward computed on a representative trajectory.}
    \label{fig:example}
    \vspace{-3mm}
\end{figure}

\vspace{-3mm}
\subsection{Out-of-span emissions}
\label{subsec:coverage}
\vspace{-1mm}

For a span's first reply, $s_\mathrm{prev}=s_0$, while silence scores $s_0/S=0.125$. Since judge scores are integers on $[0,S]$, any positive verdict improves upon silence. With $\phi=0.5$, the temporal factor in \eqref{eq:marginal} remains in $[0.5,1]$, so later responses reduce the gain but do not make speaking a loss. Only $r_{\mathrm{in\_span}}$ penalizes emissions outside annotated spans. Because it is weighted by $0.5$ and normalized by the $M$ out-of-span turns, the marginal cost of a false emission is $\kappa=0.5/M$. By contrast, perfectly answering one span contributes $3(1-s_0/S)/N=2.625/N$ to the weighted PAUC reward. Thus, for $N\leq5$, the gain from one perfectly answered span can exceed the maximum total out-of-span penalty. More generally, because span boundaries are unknown during streaming, each emission is a decision under uncertainty. Let $\rho$ denote the expected reward from hitting an unanswered span. An emission is favorable when its success probability exceeds $\kappa/(\rho+\kappa)$. Since the available reward per span scales as $1/N$ while $\kappa\propto1/M$, the incentive toward false emissions grows approximately with $M/N$. Count-insensitive rewards are therefore least conservative when events are sparse.

\begin{figure}
    \centering
    \includegraphics[width=0.48\textwidth]{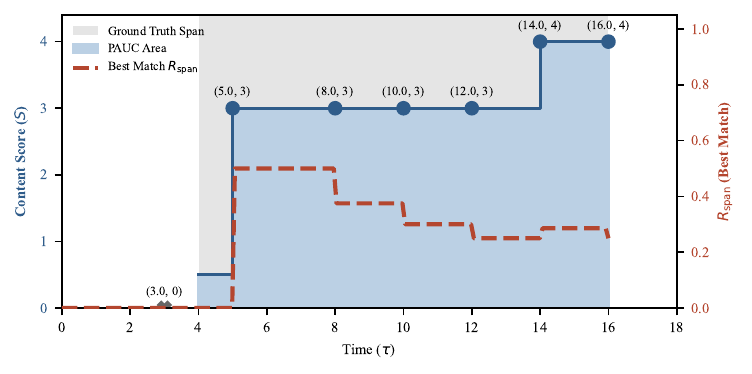}
    \vspace{-9mm}
    \caption{\footnotesize Reward response on one trajectory. Four consecutive replies scoring 3 leave $r_\mathrm{PAUC}$ unchanged, since none improves on the score in force, whereas $R_\mathrm{span}$ falls with each and rises only at $\tau{=}14$, where a higher-scoring reply justifies its cost.}
    \vspace{-5mm}
    \label{fig:reward_response}
\end{figure}

\vspace{-3mm}
\subsection{Repeated emissions within a span}
\label{subsec:retry}
\vspace{-1mm}

For subsequent replies within the same span, $s_\mathrm{prev}$ is the score currently in force. If a new reply receives the same score, \eqref{eq:marginal} evaluates to zero. Hence, PAUC itself imposes no cost for retrying, while a better-scored reply can still increase reward. When content scores exhibit variation across separately judged responses, another attempt can therefore have positive expected value whenever the current score is sufficiently low.

The auxiliary repetition terms only partially address this behavior. Because $r_\mathrm{rep}$ is averaged over emissions, adding another reply with similar repetitiveness need not substantially worsen the normalized penalty. Likewise, $r_\mathrm{pfx}$ targets lexical prefix repetition and can miss semantically redundant rephrasings. Figure~\ref{fig:reward_response} shows this directly: four identically scored replies leave $r_\mathrm{PAUC}$ unchanged, while $R_\mathrm{span}$ declines with each.
\vspace{-4mm}
\section{Emission-Aware Reward Design}
\vspace{-2mm}
\label{sec:reward}

Section~\ref{sec:analysis} suggests a general design requirement: the marginal cost of speaking should depend directly on the number of responses produced by the policy. Reweighting count-insensitive terms alone cannot guarantee this property. A precision--recall formulation naturally provides it because its
prediction-side denominator grows with model output.

Building on count-sensitive turn-level F1 reward design~\cite{li2026streampro}, we propose an \emph{emission-aware F1 reward} for interval-valued annotations. Unlike point-timestamp supervision, interval-valued targets introduce an assignment problem because multiple emissions may fall within the same span and overlapping spans may compete for the same prediction. We optimize
\begin{equation}
\begin{split}
R &= w_\mathrm{fmt}R_\mathrm{fmt}
   + w_\mathrm{span}R_\mathrm{span}, \\
R_\mathrm{span}
&=
\frac{(1+\beta^2)
\sum_{i\in\mathcal{A}}c_i\,w(\tau_i)}
{N_\mathrm{match}
+\gamma N_\mathrm{out}
+\beta^2N_\mathrm{gt}}.
\end{split}
\label{eq:reward4}
\end{equation}

Here, $R_\mathrm{fmt}$ rewards well-formed output at each decision turn, $\mathcal{A}$ is the set of matched emission–span pairs, $N_\mathrm{match} = |\mathcal{A}|$ is the number of matched emissions, $N_\mathrm{out}$ the number of unmatched ones, and $N_\mathrm{gt}$ the number of reachable ground-truth spans. We write $N_\mathrm{pred} = N_\mathrm{match} + N_\mathrm{out}$ for the total emissions. Each matched emission receives a content score $c_i\in[0,1]$ from an LLM judge and temporal weight $w(\tau_i)$. When $\gamma = \beta = 1$, the denominator becomes $N_\mathrm{pred} + N_\mathrm{gt}$, recovering the standard F1 count structure. Figure~\ref{fig:example} works through a representative trajectory.

\vspace{-3mm}
\subsection{Matching interval-valued annotations}
\label{subsec:matching}
\vspace{-1mm}

Interval-valued supervision introduces an assignment problem absent from point-based annotation. Multiple emissions may lie within a single span,
while overlapping spans can make one emission eligible for several
spans. Crediting every eligible pair would again reward
over-generation, whereas greedy matching can consume an emission needed
to cover another reachable span. We therefore use one-to-one
maximum-weight bipartite matching, allowing each emission and each gold
span to participate in at most one credited pair.

Human annotations of temporal event boundaries are also inherently
imprecise. A strict binary membership test would assign no credit to an
otherwise correct response arriving just outside the annotated boundary.
We therefore use a continuous decay around each span. For a span
$[a,b]$, let $D=\min(W,b-a)$, where $W$ caps the ramp width:
\begin{equation}
w(t)=
\begin{cases}
\max\bigl(0,\;1-\tfrac{a-t}{D}\bigr),
& t<a, \\[2pt]
\max\bigl(\alpha,\;1-\tfrac{(1-\alpha)(t-a)}{D}\bigr),
& a\leq t\leq b, \\[2pt]
\max\bigl(0,\;\alpha-\tfrac{t-b}{D}\bigr),
& t>b .
\end{cases}
\label{eq:membership}
\end{equation}
\noindent Responses near a boundary therefore receive partial credit rather than
experiencing an abrupt reward transition. The parameter $\alpha$
controls the intra-span discount for later responses; we set
$\alpha=1$, leaving the reward flat in the annotated span.

\vspace{-3mm}
\subsection{Pricing emissions and missed spans}
\label{subsec:gamma}
\vspace{-1mm}

The parameters $\gamma$ and $\beta$ control the two complementary errors. The coefficient $\gamma$ increases the cost of unmatched emissions, while $\beta$ controls the relative emphasis placed on covering ground-truth spans. Increasing $\gamma$ therefore favors a more conservative policy, whereas increasing $\beta$ places greater emphasis on recall. Exclusive matching also directly prices repeated responses. If multiple emissions compete for the same span, at most one can receive credit for that target; the others contribute to $N_\mathrm{out}$ unless they are matched elsewhere. The matching retains the highest-value, or best-match assignment, allowing a genuinely improved response to replace a weaker one while charging for unnecessary retries. The reward therefore specifies not how often the model should speak, but how much value a new response must add.

\vspace{-5mm}
\section{Experimental Results}
\label{sec:exp}
\vspace{-3mm}

\noindent\textbf{Setup.} The MMDuet2 SFT checkpoint is not publicly released, so we reproduce it from the published recipe with Qwen2.5-VL-3B~\cite{Bai2025Qwen25VLTR} initialization. From this checkpoint we run GRPO~\cite{deepseek-math} with VeRL~\cite{Sheng2025HybridFlow} and SGLang on MMDuet2's RL data for one epoch, changing only the reward; the in-reward judge is Qwen2.5-32B at temperature 0, and we set $w_\mathrm{fmt}{=}0.1$ and $w_\mathrm{span}{=}1$. We report two configurations: $\gamma{=}1.5$, which charges out-of-span emissions more than in-span ones, and $\beta{=}1.5$, an $F_\beta$ reward with soft span membership (Sec.~\ref{sec:reward}). The baseline is the released MMDuet2 checkpoint~\cite{wang2026mmduet2}. We also run the released MMDuet~\cite{wang2025mmduet} and VideoLLM-Online~\cite{videollm-online}. We evaluate on the visual subset of OmniPro~\cite{omnipro2026} and on ProactiveVideoQA~\cite{wang2025proactivevideoqa}. For both, timing precision, recall and F1 use one-to-one matching between replies and gold answers; $N_{pred}/N_{gt}$ counts replies per gold answer.

\noindent\textbf{Results on OmniPro.} OmniPro annotates each answer as a point trigger; a reply matches a trigger within 3\,s. To score what a reply says as well as when, we define F1$_c = 2S_c/(N_{pred}+N_{gt})$, where $S_c$ sums a GPT-4.1 content score (0, 0.5 or 1, against the trigger's reference answer) over matched pairs; timing F1 is the case where every match scores 1. Table~\ref{tab:omnipro} shows two opposite ways a proactive model can fail. VideoLLM-Online almost never replies and recovers 1.0\% of triggers. MMDuet replies 5.39 times per trigger and reaches the highest recall (0.738), but only 13.7\% of its replies match a trigger. Our policies sit between these extremes. The $\beta$ policy replies 1.63 times per trigger, 2.5$\times$ fewer than MMDuet2 and 3.3$\times$ fewer than MMDuet, and achieves the highest timing F1 (0.396) and content F1$_c$ (0.154) of all models. The $\gamma$ policy replies 0.71 times per trigger with the highest precision (0.418), above even the nearly silent VideoLLM-Online (0.395). F1$_c$ improves along with timing F1, so the remaining replies are both better placed and more often correct. MMDuet2 reaches higher recall than either policy by generating more responses to cover each trigger, while $\beta$ retains 80\% of that recall with 40\% of the replies.

% \begin{table}[t]
% \centering
% \footnotesize
% \setlength{\tabcolsep}{3.5pt}
% \begin{tabular}{lccccc}
% \toprule
% \textbf{Model} & \textbf{Fires} $\downarrow$ & \textbf{Prec.} $\uparrow$
%   & \textbf{Rec.} $\uparrow$ & \textbf{F1} $\uparrow$ & \textbf{F/T} $\downarrow$ \\
% \midrule
% VideoLLM-Online~\cite{videollm-online} & 43  & - & - & -  & - \\
% MMDuet~\cite{wang2025mmduet}                  & 16009 & 0.142 & \textbf{0.875} & 0.168 & 1.51 \\
% MMDuet2~\cite{wang2026mmduet2}                & 6961 & 0.261 & 0.665 & 0.259 & 1.59 \\
% \textbf{Ours ($\beta=1.5$)}              & 2804  &  \textbf{0.540}   &   0.379 & \textbf{0.396} & - \\
% \textbf{Ours ($\gamma=1.5$)}           & \textbf{1216} & \underline{0.448} & 0.309 & 0.346
%                                       & \textbf{1.02} \\
% \bottomrule
% \end{tabular}
% \vspace{-3mm}
% \caption{\footnotesize Visual subset of OmniPro ($n{=}432$, 1723 triggers, tolerance 3.0\,s). All columns use exclusive matching: each gold trigger is credited to at most one emission. F/T is emissions per covered trigger.}
% \label{tab:omnipro}
% \vspace{-5mm}
% \end{table}
% %\vspace{-3mm}

\begin{table}[t]
\centering
\footnotesize
\setlength{\tabcolsep}{3.5pt}
\begin{tabular}{lccccc}
\toprule
\textbf{Model} & $\mathbf{N_{pred}/N_{gt}}$ & \textbf{Prec.}$\uparrow$
  & \textbf{Rec.}$\uparrow$ & \textbf{F1}$\uparrow$ & \textbf{F1$_{c}$}$\uparrow$ \\
\midrule
VideoLLM-Online~\cite{videollm-online} & 0.02 & 0.395 & 0.010 & 0.019 & 0.003 \\
MMDuet~\cite{wang2025mmduet}           & 5.39 & 0.137 & \textbf{0.738} & 0.231 & 0.058 \\
MMDuet2~\cite{wang2026mmduet2}         & 4.04 & 0.161 & \underline{0.652} & 0.259 & 0.103 \\
\textbf{Ours ($\beta=1.5$)}            & 1.63 & \underline{0.320} & 0.521 & \textbf{0.396} & \textbf{0.154} \\
\textbf{Ours ($\gamma=1.5$)}           & 0.71 & \textbf{0.418} & 0.295 & \underline{0.346} & \underline{0.138} \\
\bottomrule
\end{tabular}
\vspace{-3mm}
\caption{\footnotesize Visual subset of OmniPro ($n{=}432$, $N_{gt}{=}1723$ triggers).}
\label{tab:omnipro}
\vspace{-7mm}
\end{table}

\begin{table}[t]
\centering
\scriptsize
\setlength{\tabcolsep}{5pt}
\setlength{\aboverulesep}{0pt}
\setlength{\belowrulesep}{0pt}
\setlength{\extrarowheight}{1pt}
\begin{tabular}{@{}cl cccc@{}}
\toprule
& \textbf{Model} & \textbf{PAUC}$\uparrow$ & \textbf{Rep\%}$\downarrow$
  & $\mathbf{N_{pred}/N_{gt}}$ & \textbf{F1}$\uparrow$\\
\midrule
\multirow{5}{*}{\rotatebox[origin=c]{90}{\textbf{EGO}}}
 & VideoLLM-Online & 25.3 & 18.5 & 0.73  & 0.024 \\
 & MMDuet          & 43.6 & 77.2 & 16.86 & 0.092 \\
 & MMDuet2         & 32.1 & 9.5  & 0.54  & 0.248 \\
 & \textbf{Ours ($\beta=1.5$)}  & \textbf{46.0} & 13.4 & 1.72 & \underline{0.418} \\
 & \textbf{Ours ($\gamma=1.5$)} & 37.7 & \textbf{2.5} & 0.79 & \textbf{0.419} \\
\cmidrule(l){2-6}
\multirow{5}{*}{\rotatebox[origin=c]{90}{\textbf{TV}}}
 & VideoLLM-Online & 20.0 & 32.0 & 5.69  & 0.100 \\
 & MMDuet          & 22.5 & 61.0 & 20.62 & 0.078 \\
 & MMDuet2         & 45.6 & 1.3  & 2.37  & \textbf{0.358} \\
 & \textbf{Ours ($\beta=1.5$)}  & \textbf{47.0} & 3.0 & 2.92 & \underline{0.350} \\
 & \textbf{Ours ($\gamma=1.5$)} & 38.6 & \textbf{0.0} & 1.63 & 0.316 \\
\cmidrule(l){2-6}
\multirow{5}{*}{\rotatebox[origin=c]{90}{\textbf{WEB}}}
 & VideoLLM-Online & 25.2 & 3.3 & 0.07 & 0.097 \\
 & MMDuet          & 41.3 & 9.3 & 0.79 & \underline{0.703} \\
 & MMDuet2         & \textbf{54.4} & 4.4 & 1.25 & \textbf{0.718} \\
 & \textbf{Ours ($\beta=1.5$)}  & 45.4 & 2.3 & 0.75 & 0.680 \\
 & \textbf{Ours ($\gamma=1.5$)} & 36.5 & \textbf{0.7} & 0.38 & 0.495 \\
\cmidrule(l){2-6}
\multirow{5}{*}{\rotatebox[origin=c]{90}{\textbf{VAD}}}
 & VideoLLM-Online & 25.0 & --   & 0.07  & 0.000 \\
 & MMDuet          & 23.1 & 43.0 & 29.11 & 0.052 \\
 & MMDuet2         & \textbf{29.4} & 12.7 & 1.68 & \underline{0.188} \\
 & \textbf{Ours ($\beta=1.5$)}  & 27.7 & 15.5 & 1.64 & \textbf{0.191} \\
 & \textbf{Ours ($\gamma=1.5$)} & 27.8 & \textbf{0.0} & 0.75 & 0.182 \\
\bottomrule
\end{tabular}
\vspace{-3mm}
\caption{\footnotesize ProactiveVideoQA ($N_{gt}$: EGO 1575, TV 500, WEB 1328, VAD 107). PAUC floors at 25 for a silent model. F1 is timing-only: a reply matches a gold turn if inside its reply span, one reply per turn; with $r{=}N_{pred}/N_{gt}$, precision $=\mathrm{F1}(1{+}1/r)/2$ and recall $=\mathrm{F1}(1{+}r)/2$.}
\label{tab:pvqa}
\vspace{-6mm}
\end{table}

\noindent\textbf{Results on ProactiveVideoQA.} We report PAUC and repetition (Rep\%) with MMDuet2's released scorer and a GPT-4.1 judge, where Rep\% is the fraction of repetitive replies, averaged over spans. PAUC integrates the best answer given within each span and, as Sec.~\ref{sec:analysis} argues, cannot observe reply volume. Table~\ref{tab:pvqa} shows this with the baselines. On EGO, MMDuet ranks second on PAUC (43.6) while replying 16.9 times per gold answer, yet second-lowest on F1. On VAD, its 29 replies per answer score 23.1 PAUC, below the 25 a silent model receives. Our reward does not simply suppress replies. On EGO, where MMDuet2 replies only 0.54 times per gold answer and stays silent on 39.6\% of videos, our policies reply 0.79--1.72 times per answer and raise F1 from 0.248 to 0.419, the highest of all models; $\gamma$ improves every column over MMDuet2 while speaking more, and $\beta$ attains the highest PAUC on both EGO (46.0) and TV (47.0). On TV and VAD, $\beta$ is comparable to MMDuet2 in F1; on WEB it trails: WEB's gold spans cover 81\% of video time, so a reply at a random moment already lands in a span 81\% of the time, and dense firing costs almost nothing. Interval annotation at this density leaves little room for a reward that prices replies. Hence, OmniPro's point triggers serve as our primary test.

\vspace{-3mm}
\section{Conclusions}
\label{sec:conclusion}
\vspace{-3mm}
We show that over-emission in proactive video assistants can arise from reward design rather than failed optimization. We identify a general failure mode in PAUC-based objectives, where additional predictions can improve reward without a count-dependent cost, and address it with an emission-aware F1 reward using exclusive span--emission matching. On OmniPro, our policy reduces emissions by 82.5\% while raising precision from 0.161 to 0.418 and F1 from 0.259 to 0.346. Limitations include intra-span timing and manually chosen $\gamma$ and $\beta$, which future work can derive from deployment-specific costs.
\vspace{-4mm}
\section{Acknowledgement}
\vspace{-3mm}
This work was supported by the Air Force Research Laboratory/Information and Spectrum Warfare Directorate (AFRL/RF), Rome, NY under the 6.2 Program Planar Bridge. Any statements or claims made in this publication are solely those of the authors and do not represent the views or policies of the United States Government.

\newpage
\section{Compliance with Ethical Standards}
This study uses existing publicly available benchmark datasets and
pretrained models and does not involve the collection of new human
subject data. Therefore, no additional ethical approval was required.

\small
\bibliographystyle{IEEEbib}
\bibliography{icassp_2027}

\end{document}